\documentclass[10pt, a4paper, dvipsnames]{article}

\usepackage[]{lrec-coling2024} 

\usepackage{xstring}

\usepackage{xcolor}
\usepackage{multirow}
\usepackage{booktabs}

\title{Event Extraction in Basque: \\ Typologically  motivated Cross-Lingual Transfer-Learning Analysis}

\name{Mikel Zubillaga \quad Oscar Sainz \quad Ainara Estarrona \smallskip \\ \medskip {\bf \large Oier Lopez de Lacalle \quad Eneko Agirre}} 

\address{HiTZ Basque Center for Language Technology - Ixa NLP Group \\
        University of the Basque Country UPV/EHU \\
        \{mikel.zubillaga, oscar.sainz, ainara.estarrona, oier.lopezdelacalle, e.agirre\}@ehu.eus\\}

\abstract{
Cross-lingual transfer-learning is widely used in Event Extraction for low-resource languages and involves a Multilingual Language Model that is trained in a source language and applied to the target language. 
This paper studies whether the typological similarity between source and target languages impacts the performance of cross-lingual transfer, an under-explored topic. 
We first focus on Basque as the target language, which is an ideal target language because it is typologically different from surrounding languages.  Our experiments on three Event Extraction tasks show that the shared linguistic characteristic between source and target languages does have an impact on transfer quality.
Further analysis of 72 language pairs reveals that for tasks that involve token classification such as entity and event trigger identification, common writing script and morphological features produce higher quality cross-lingual transfer. In contrast, for tasks involving structural prediction like argument extraction, common word order is the most relevant feature. In addition, we show that when increasing the training size, not all the languages scale in the same way in the cross-lingual setting. To perform the experiments we introduce EusIE, an event extraction dataset for Basque, which follows the Multilingual Event Extraction dataset (MEE). The dataset and code are publicly available. 
 \\ \newline  }

\begin{document}

\maketitleabstract

\section{Introduction}

\begin{table*}
    \centering
    \begin{tabular}{l|l}
        \toprule
        Sentence & Peter, the CEO of XYZ company, got married in Brazil. \\
        \midrule
        Entities & {\color{RoyalPurple}[}Peter{\color{RoyalPurple}]\textsubscript{PER}}, the {\color{RoyalPurple}[}CEO{\color{RoyalPurple}]\textsubscript{JOB}} of {\color{RoyalPurple}[}XYZ{\color{RoyalPurple}]\textsubscript{ORG}} company, got married in {\color{RoyalPurple}[}Brazil{\color{RoyalPurple}]\textsubscript{LOC}}. \\
        \rule{0pt}{2ex}Events & Peter, the CEO of XYZ company, {\color{RubineRed}[}got married{\color{RubineRed}]\textsubscript{Life.Marry}} in Brazil. \\
        \rule{0pt}{2ex}Arguments & {\color{Cerulean}[}Peter{\color{Cerulean}]\textsubscript{Person-arg}}, the CEO of XYZ company, {\color{RubineRed}\$\$\$}got married{\color{RubineRed}\$\$\$} in {\color{Cerulean}[}Brazil{\color{Cerulean}]\textsubscript{Place-arg}}.\\
        \bottomrule
    \end{tabular}
    \caption{A sample sentence annotated according to the three tasks related to Event Extraction. }
    \label{tab:ArgAdib}
\end{table*}


Event Extraction (EE) is one of the fundamental tasks in Information Extraction (IE) and aims to extract event mentions and their arguments (i.e., participants) from text. Typically, EE involves the identification of trigger words (e.g. \emph{married}, \emph{the attack}) that denote a mention of an event or action. In parallel, the entities in the sentence are extracted. In a final step, once the event is known, the participants that take part in the event are identified in the context. Table~\ref{tab:ArgAdib} shows a complete example of an event extraction pipeline, in which we show the entities, event triggers, and the arguments of the corresponding event mention.  Due to the complexity involved, and the high interest in the task, EE has been historically one of the most relevant tasks in the field of Information Extraction. 

Information Extraction tasks in general and EE in particular pose significant challenges~\citep{grishman_2019}, as it is a complex task that demands humans to meticulously follow complicated guidelines, often riddled with numerous exceptions. To tackle this challenge, the conventional approach is to train computational models with large amounts of annotated examples. Obtaining the examples entails extensive manual effort by domain experts, making it impractical for situations with limited resources, especially for low-resource languages. 

Due to the recent advances in Natural Language Processing~\cite{10.1145/3605943}, Large Language Models (LLMs) are capable of transferring knowledge across languages, i.e. training in one language and performing inferences in another. This is referred to as Cross-Lingual Transfer Learning and has represented a significant advancement for languages other than English, as it allows to obtain EE models using data from high-resource languages (primarily English). The success of this approach allowed to develop ambitious programs, such as BETTER~\cite{mckinnon-rubino-2022-iarpa}, where English data is provided for training while the models are tested on other languages. While the proposed task was interesting and really challenging, it only uses English data for training. In recent work, ~\citet{pouran-ben-veyseh-etal-2022-mee} developed MEE (Multilingual Event Extraction), an extension of the well-known ACE 2005~\citeplanguageresource{ACE} dataset to 8 languages.  

One limitation of current EE research is the under-exploration of non-English languages, due mainly to the lack of high-quality multilingual dataset. MEE allows for such kind of research, and we thus use MEE to explore whether the typology of target and source languages impacts cross-lingual transfer capabilities. In particular, we study what would be the best transfer choice to develop an IE system for a language with no training data. 

In order to increase the typological diversity of languages in MEE, we added Basque, a language isolate with no known related language. The Basque language has a particularly interesting set of features, very different from the surrounding languages, making it an interesting candidate to study which features of languages affect the quality of cross-lingual transfer.  We thus annotated an evaluation set for Basque, \textbf{the first Event Extraction benchmark for the language}\footnote{\href{https://huggingface.co/datasets/HiTZ/EusIE}{https://huggingface.co/datasets/HiTZ/EusIE}}. We follow the same procedure as in ~\citep{pouran-ben-veyseh-etal-2022-mee} when collecting and annotating examples, with the exception that we used expert annotators in contrast to crowd-working services. Due to the high cost, we only annotated evaluation data.

In our experiments, we first explored which linguistic characteristics affect positively and negatively when we evaluate in Basque. We thus trained several models on each MEE language under the same conditions and evaluated them in Basque. Further, we extend this analysis to all the languages and measure systematically how the language typology affects cross-lingual transfer. The code of the experiments is publicly available\footnote{\href{https://github.com/MikelZubi/GrAL}{https://github.com/MikelZubi/GrAL}}
%
%
The results show that the transfer quality depends on the shared linguistic characteristic between source and target language, but varies across each of the tasks. Further analysis reveals that for tasks involving token classification (i.e. entity and trigger identification) sharing writing script shows higher cross-lingual transfer benefit. In contrast, when structural understanding is involved (e.g. argument extraction) word order matters the most.  

\section{Related Work}

\paragraph{Event Extraction.} 



Early methods addressed the task by defining human-crafted features and applying rules~\cite{ji-grishman-2008-refining, gupta-ji-2009-predicting, hong-etal-2011-using, li-etal-2013-joint}. These methods were replaced by deep learning approaches~\cite{chen-etal-2015-event, feng-etal-2016-language, liu-etal-2018-jointly} in the last decade. Soon, sequence labeling became the standard approach for EE~\cite{nguyen-etal-2016-joint-event, chen-etal-2018-collective, araki-mitamura-2018-open, ding-etal-2019-event, lin-etal-2020-joint, guzman-nateras-etal-2022-cross}. With the development of pre-trained Large Language Models, several works reformulated the task into language understanding tasks such as Question Answering~\cite{du-cardie-2020-event, li-etal-2020-event, liu-etal-2020-event, wei-etal-2021-trigger, sheng-etal-2021-casee, zhou-etal-2022-multi} and Textual Entailment~\cite{sainz-etal-2022-textual, sainz-etal-2022-zs4ie} to benefit from the implicit knowledge and capabilities encoded the model. 
Recently, with the increasing popularity of generative models, works based on conditional generation have also been proposed~\cite{xiangyu-etal-2021-capturing, lu-etal-2021-text2event, li-etal-2021-document, hsu-etal-2022-degree, zeng-etal-2022-ea2e, li-etal-2022-kipt, liu-etal-2022-dynamic, huang-etal-2022-multilingual-generative, du-etal-2022-dynamic}. Lastly, multi-task instruction-based models have been applied to perform several tasks together, including event extraction ~\cite{DBLP:journals/corr/abs-2304-08085, sainz2023gollie}. 

\paragraph{Cross-lingual approaches for IE}
Pre-trained LLMs allowed a simplified approach to cross-lingual IE with state-of-the-art performance
~\cite{XLMRoberta,XLM,conneau-etal-2020-unsupervised}. Previously, state-of-the-art consisted of using parallel data to project labels from one language to the other ~\cite{agerri-etal-2018-building}. Related to this, the improvement of machine translation and alignment models allowed effective augmentation of training examples
~\cite{garcia-ferrero-etal-2022-model,li2021crosslingual, Lou2022}. 
In this paper, we choose to use a multilingual sequence labeling approach to efficiently analyze the characteristics of cross-lingual transfer learning.

\begin{table*}
    \centering
    \begin{tabular}{l|l|l}
        \toprule
         Entities & \multicolumn{2}{l}{PER, ORG, GPE, LOC, FAC, VEH, WEA, CRIME, TIME, MON, POS, \textit{OBJ}} \\
         \midrule
         \multirow{16}{*}{Events} & Life:Be-Born & Person, Time, Place \\
          & Life:Marry & Person, Time, Place \\
          & Life:Divorce & Person, Time, Place \\
          & Life:Injure & Agent, Victim, Instrument, Time, Place \\
          & Life:Die & Agent, Victim, Instrument, Time, Place \\
          & Movement:Transport & Agent, Artifact, Vehicle, Price, Origin, Destination, Time \\
          & Transaction:Transfer-Ownership & Buyer, Seller, Beneficiary, Price, Artifact, Time, Place \\
          & Transaction:Transfer-Money & Giver, Recipient, Beneficiary, Money, Time, Place \\
          & Business:Start-Organization & Agent, Organization, Time, Place \\
          & Conflict:Attack & Attacker, Target, Instrument, Time, Place \\
          & Conflict:Demonstrate & Entity, Time, Place \\
          & Contact:Meet & Entity, Time, Place \\
          & Contact:Phone-Write & Entity, Time \\
          & Personnel:Start-Position & Person, Entity, Position, Time, Place \\
          & Personnel:End-Position & Person, Entity, Position, Time, Place \\
          & Justice:Arrest-Jail & Person, Agent, Crime, Time, Place \\
         \bottomrule
    \end{tabular}
    \caption{Annotation schema used to annotate EusIE. The schema is the same as the one used by MEE and is based on ACE 2005. Except for the label \textit{OBJ} that does not exist on MEE, and therefore, it is not used for evaluation.}
    \label{tab:annotation_schema}
\end{table*}

\paragraph{Existing datasets} for Event Extraction are mostly available only for English, such as CySecED~\cite{man-duc-trong-etal-2020-introducing}, CASIE~\cite{casie}, LitBank~\cite{sims-etal-2019-literary}, MAVEN~\cite{wang-etal-2020-maven}, RAMS~\cite{ebner-etal-2020-multi} and, WikiEvents~\cite{li-etal-2021-document} among others. Additionally, there are a few multilingual EE datasets like ACE 2005~\citeplanguageresource{ACE} and more recently BETTER~\cite{mckinnon-rubino-2022-iarpa} and MEE~\cite{pouran-ben-veyseh-etal-2022-mee}. ACE 2005 and BETTER include only English training data. MEE contains annotated train and evaluation datasets in eight languages. In this work, we follow ACE 2005 and MEE guidelines, and annotate a Basque Event Extraction dataset to perform our experiments.

\section{EusIE: Basque Event Extraction}

In this section, we present EusIE (\textbf{Eus}karazko \textbf{I}nformazio-\textbf{E}rauzketa)\footnote{\textit{Basque Information-Extraction} in Basque.}, which is the first EE dataset for Basque. We decided to extend the Multilingual Event Extraction (MEE) dataset (~\cite{pouran-ben-veyseh-etal-2022-mee})  by following the well-known  ACE05~\citeplanguageresource{ACE} ontology. The MEE dataset covers 8 diverse languages that we use in our experiments in conjunction with Basque. 

Although the dataset creation process followed similar steps to MEE, few modifications were implemented. On one hand, due to the difficulty of finding Basque-speaking crowd workers, two native experts annotated the dataset. On the other hand, due to our small budget, we limit the annotation to the development and test splits. This way we provide quality over quantity. In the following sections, we describe the data collection, filtering, and annotation process.

\subsection{Data collection}

We collect the initial set of documents from a snapshot of Basque Wikipedia\footnote{The downloaded snapshot was from October 10th, 2022. \url{https://dumps.wikimedia.org/other/cirrussearch/current/}}. From the initial set, we select the documents related to events (\textit{Gertaerak} category) that were labeled as part of the following topics: Economy (\textit{Ekonomia}), Politics (\textit{Politika}), Technology (\textit{Teknologia}), Natural Disasters (\textit{Hondamen Naturalak}), Military (\textit{Militarrak}) and Crimes (\textit{Krimenak}). We keep the same topics as the original MEE to avoid domain shifts.

After collecting the documents, we removed the markup from the documents using WikiExtractor~\cite{Wikiextractor2015}. Additionally, section titles and other structural information were removed too. We split the documents into sentences, and, we tokenized them using IXA-pipes~\cite{agerri-etal-2014-ixa} an NLP toolkit designed for Basque. Similar to MEE and RAMS~\cite{ebner-etal-2020-multi}, 
we grouped 5 sentences to form an annotation \textbf{segment}.
The segment is our annotation boundary, and thus the relations between the events and arguments can occur within the sentence as well as cross sentences, but always inside the segments. 

\subsection{Annotation schema}

The annotation schema used to annotate the dataset is shown in Table~\ref{tab:annotation_schema}. We adapted the schema used in MEE to include entity types that could potentially be argument candidates for the defined events. We included \texttt{OBJ} to categorize entities that are candidates for the \texttt{Artifact} role. We do not have examples for the \texttt{NUM} entity type, as all numerical mentions could be labeled either with \texttt{DATE} or \texttt{MON}.

\subsection{Annotation}

\begin{table}
    \centering
    \resizebox{1.\linewidth}{!}{\begin{tabular}{l|rrrr}
         \toprule
         Language & Tokens & Entities & Events & Arguments \\
         \midrule
         English & 123 & 14.66 & 1.36 & 1.04 \\
         Spanish & 112 & 14.69 & 1.85 & 0.24 \\
         Portuguese & 102 & 16.98 & 1.30 & 8.21 \\
         Polish & 108 & 14.06 & 2.42 & 0.76 \\
         Turkish & 117 & 8.59 & 1.87 & 0.31 \\
         Hindi & 98 & 12.53 & 1.21 & 1.41 \\
         Japanese & 99 & 12.78 & 1.44 & 2.27 \\
         Korean & 103 & 8.34 & 0.75 & 1.16 \\
         \midrule
         \textit{Mean} & 108 & 12.83 & 1.53 & 1.93 \\
         \midrule
         Basque & 94 & 16.58 & 2.17 & 4.49 \\
         \bottomrule
    \end{tabular}}
    \caption{Average statistic per segment for each language.}
    \label{tab:statistics}
\end{table}

We annotated a total of 300 segments (1500 sentences) and divided them into 150 for development and the rest of 150 for testing. That is, we annotated a similar amount of segments provided in the evaluation partition of the MEE dataset. 
The annotator was provided with the Inception~\cite{klie-etal-2018-inception} annotation tool and the ACE 2005 guidelines\footnote{\url{https://www.ldc.upenn.edu/sites/www.ldc.upenn.edu/files/english-events-guidelines-v5.4.3.pdf}}. Statistics of the annotation process are shown in Table~\ref{tab:statistics}. Overall, the annotated segments contain substantially more annotations than the average.

To ensure annotation quality, we asked a second expert to annotate a portion of the data, 35 segments and computed the Cohen's $\kappa$ between both annotators. For span annotations which include entities and event triggers, the annotators obtained an agreement of 0.94, and, for the arguments, the annotators obtained an agreement of 0.92.  The obtained agreement is indicative of the quality of the expert annotators and the annotation guidelines.

The annotated data was converted from WebAnno 3.0 to a JSON format introduced by ~\citet{lin-etal-2020-joint} for simplicity. Once converted, the data was split into development and test ensuring that segments from the same original document remain on the same partition.

\section{Experimental Setup}

In this work, we explore the cross-lingual capabilities of the multilingual Language Models in EE for the Basque language.  We deploy the aforementioned MEE and EusIE datasets. Typically, EE is a sequence of three tasks that are evaluated as a pipeline, reporting the final F1 results.  As we want to compare the transfer qualities of each language empirically in each of the pipelined steps, we reported the F1 scores for each task (entity, event, and argument extraction).  All the tasks are evaluated independently using gold annotations from the previous step in the pipeline~\footnote{E.g. we used gold event triggers when detecting the arguments.}. Additionally, three different runs are executed for each configuration in order to provide average and deviation scores.

We organize the experiments in three main parts. First, we try to replicate the in-language experiments performed in ~\citep{pouran-ben-veyseh-etal-2022-mee}, where models are evaluated and trained in the same language. We use their baseline approach (called \textit{Pipeline} in the paper), trained and evaluated in the original data splits provided by the authors.
Next,  we run the cross-lingual experiments, where we train models in one language and evaluate them in Basque (EusIE). For fair comparisons across languages, all languages have the same number of train examples, i.e. we discard examples in the languages with most training data. More details in Section~\ref{sec:remove_data}.
Finally, we run an analysis of linguistic features to gain insight into what makes a language good for cross-lingual transfer-learning. We will categorize each language based on its typological features, and perform all-vs-all experiments to analyze the impact of those features.

\subsection{Model}
As mentioned above, EE is typically divided into 
entity detection, event detection, and, event argument extraction. 
Therefore, we train three models, one per each task. 
As shown in Table~\ref{tab:ArgAdib}, we formulated all the tasks as sequence labeling problems. Both, entity and event detection tasks are simply formulated as predicting the label for each token in the input text. For the event argument-extraction task, however, the output must be conditioned on the event to analyze. As we indicate in the Table~\ref{tab:ArgAdib}, we surround the event trigger with markers, ''\$\$\$" in our case, and label only the corresponding arguments. The backbone language model is the base version of XLM-RoBERTa~\cite{XLMRoberta}.




Regarding the hyperparameters, we set their values based on a few preliminary experiments.  The overall best performing hyperparameters were a learning-rate of 5e\textsuperscript{-5}, 32 for the batch-size and a weight decay of 1e\textsuperscript{-3}. We run the models for 64 epochs, as we found out that the F1 score was increasing in the development even if the loss was increasing too.



\subsection{Comparable Training Size} \label{sec:remove_data}
In order to compare the cross-lingual transfer capability of the languages in a comparable manner (results reported in Section~\ref{sec:results_eusie}), we try to control the size of the training data. We thus equalize the amount of training data for all the languages. That is, we remove training examples from larger languages until all the languages contain the same amount of annotations. Note that we performed the under-sampling by counting annotations and not the number of segments, as the latter could lead to a different number of annotations per language. We also add examples from development and testing in order to increase the size of the training set (note that this training is uniquely used in the cross-lingual experiments where models are evaluated in Basque). As a result, for each task, each language's data was reduced to the amount of data for the language with the least amount of annotations: 12508 annotated entities, 1125 event mentions, and 1416 arguments.

\begin{table}
    \centering
    \resizebox{1.\linewidth}{!}{\begin{tabular}{l|rrr}
        \toprule
        Languages & \multicolumn{1}{c}{Entities} & \multicolumn{1}{c}{Events} & \multicolumn{1}{c}{Arguments} \\
        \midrule
        English & 80.48${\scriptstyle\pm}$\scriptsize{0.19} & \textbf{78.47}${\scriptstyle\pm}$\scriptsize{0.33} & 63.60${\scriptstyle\pm}$\scriptsize{0.09}\\
        Spanish & \textbf{84.56}${\scriptstyle\pm}$\scriptsize{0.38} & 63.86${\scriptstyle\pm}$\scriptsize{1.20} & 40.45${\scriptstyle\pm}$\scriptsize{1.91} \\
        Portuguese & 80.42${\scriptstyle\pm}$\scriptsize{0.32} & 61.79${\scriptstyle\pm}$\scriptsize{0.63} & 68.18${\scriptstyle\pm}$\scriptsize{0.99} \\
        Polish & 81.00${\scriptstyle\pm}$\scriptsize{0.40} & 69.09${\scriptstyle\pm}$\scriptsize{0.91} & \textbf{76.25}${\scriptstyle\pm}$\scriptsize{1.26} \\
        Turkish & 70.83${\scriptstyle\pm}$\scriptsize{0.06} & 56.62${\scriptstyle\pm}$\scriptsize{0.43} & 24.08${\scriptstyle\pm}$\scriptsize{1.66} \\
        Hindi & 76.00${\scriptstyle\pm}$\scriptsize{0.41} & 48.21${\scriptstyle\pm}$\scriptsize{2.54} & 45.31${\scriptstyle\pm}$\scriptsize{1.55} \\
        Japanese & 47.43${\scriptstyle\pm}$\scriptsize{0.20} & 35.74${\scriptstyle\pm}$\scriptsize{1.92} & 52.93${\scriptstyle\pm}$\scriptsize{1.26} \\
        Korean & 71.31${\scriptstyle\pm}$\scriptsize{0.78} & 45.30${\scriptstyle\pm}$\scriptsize{0.49} & 34.71${\scriptstyle\pm}$\scriptsize{3.06} \\
        \midrule
        All & 78.63${\scriptstyle\pm}$\scriptsize{0.17} & 68.01${\scriptstyle\pm}$\scriptsize{0.27} & 59.74${\scriptstyle\pm}$\scriptsize{0.99} \\
        \bottomrule
    \end{tabular}}
    \caption{Results obtained by our model for each task and language. \emph{All} reports results obtained after training and testing with all languages together.}
    \label{tab:testO}
\end{table}

\section{Results}
In this section, we discuss the results obtained in the experiments. 
First, similar to the MEE authors, we report the results using the original splits.
Second, we report the results obtained on the EusIE benchmark. 
Finally, we explore the effect of scaling training data.


\subsection{Result on In-language Scenario}

Table~\ref{tab:testO} shows the F1 scores obtained in each language. Additionally, we included the \textit{All} row, which represents a model trained and tested using all the languages available in the dataset. We repeated each experiment 3 times and reported the average F1 score and the standard deviation for each setting. Language-wise comparisons do not show any clear pattern in which we can distinguish a particular language that overperforms the rest of the languages across all the tasks. A language that performs strongly in a specific task, shows poor performance in the other task (e.g. Spanish shows an outstanding 84.5 of F1 in entity detection, whereas it is far from top results in Argument identification). 

\begin{figure}
    \centering
    \includegraphics[width = \linewidth]{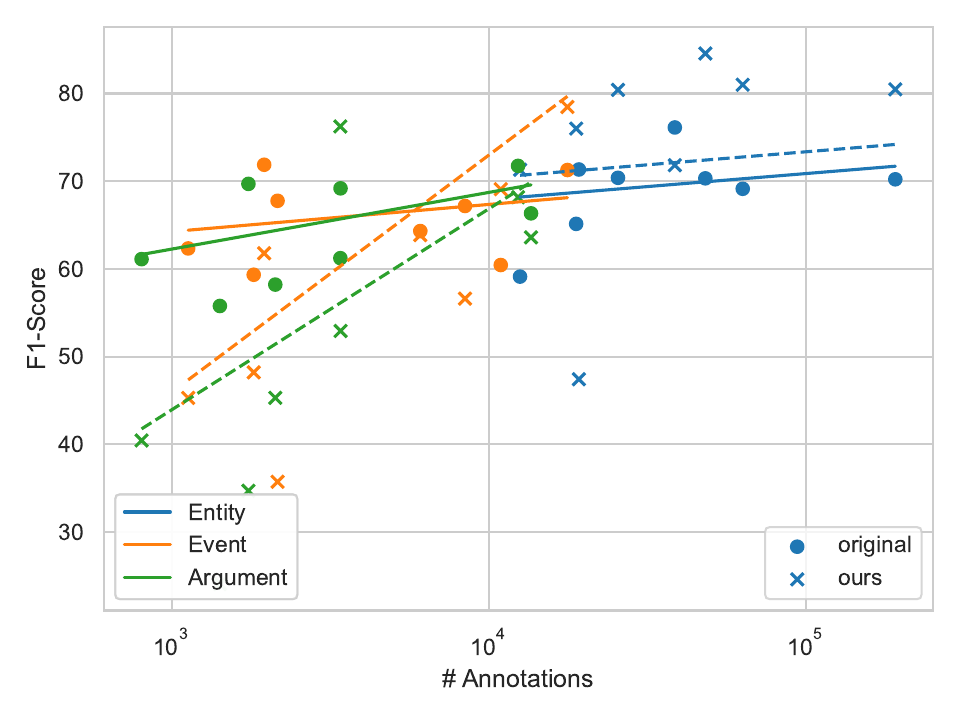}
    \caption{Comparison between our model and the original reported results based on the number of total annotations and F1-Score. Dashed lines show the linear relationship of our system and the number of annotation in training. The continuous line refers to~\citep{pouran-ben-veyseh-etal-2022-mee}.}
    \label{fig:train-correlation}
\end{figure}

It is important to note that our results substantially deviate from the ones reported in the original paper~~\citep{pouran-ben-veyseh-etal-2022-mee}\footnote{See Table~\ref{tab:comparison-mee} in Appendix~\ref{ap:results} for comparing the results with the original ones}. Discussion with the authors did not reveal any reason for this difference apart from the use of a different model.
However, we found that the results obtained by our system correlate better with the number of annotations in the training set, as shown in Figure~\ref{fig:train-correlation}. The linear correlation of our model is plotted with dashed lines, the original ones with continuous lines. In particular, for event and argument detection, in which the number of annotations is much smaller compared to entity detection, our system linearly improved when we increased the number of annotations. In the case of entities, as we have more annotations, it does not show a significant positive relationship with the number of annotations (both systems show similar behavior in this case). 

\subsection{Results on EusIE}\label{sec:results_eusie}

Table~\ref{tab:testB} shows the results for the models trained on each language when evaluated in Basque. Note that, in this experiment, all the languages have the same amount of training examples (cf. Section~\ref{sec:remove_data}). The best-performing language varies across tasks. We had hypothesized that the best results would be for Spanish, as it is an official language in the Basque Country and the contact between the two languages has been happening since Spanish became a language on its own, but that is not the case.

Regarding entity detection, results and the standard deviation show that English, Portuguese, and Polish obtain very similar results and outperform the rest of the languages, whereas Spanish, Turkish, and Hindi are close to the best results. 
The results obtained with Japanese and Korean, lag significantly behind the rest by a large margin.

We observe a similar pattern for event detection but with larger differences between languages. The best-performing language, in this case, is Polish followed by Turkish, Spanish, and English. The rest of the languages fall behind, Japanese in particular. It is important to note that the standard deviations are very high for some of the languages. 




\begin{table}
    \centering
    \resizebox{1.\linewidth}{!}{\begin{tabular}{l|rrr}
        \toprule
        Languages & \multicolumn{1}{c}{Entities} & \multicolumn{1}{c}{Events} & \multicolumn{1}{c}{Arguments} \\ 
        \midrule
        English & \textbf{59.65}${\scriptstyle\pm}$\scriptsize{1.19} & 42.65${\scriptstyle\pm}$\scriptsize{6.57} & 13.93${\scriptstyle\pm}$\scriptsize{2.06} \\
        Spanish & 56.56${\scriptstyle\pm}$\scriptsize{0.45} & 43.71${\scriptstyle\pm}$\scriptsize{2.63} & \underline{2.88}${\scriptstyle\pm}$\scriptsize{0.79} \\
        Portuguese & 59.62${\scriptstyle\pm}$\scriptsize{1.67} & 24.30${\scriptstyle\pm}$\scriptsize{2.14} & 14.02${\scriptstyle\pm}$\scriptsize{0.96}\\
        Polish & 59.48${\scriptstyle\pm}$\scriptsize{1.35} & \textbf{46.37}${\scriptstyle\pm}$\scriptsize{1.94} & 10.28${\scriptstyle\pm}$\scriptsize{0.29} \\
        Turkish & 55.72${\scriptstyle\pm}$\scriptsize{2.49} & 44.46${\scriptstyle\pm}$\scriptsize{2.73} & 14.84${\scriptstyle\pm}$\scriptsize{3.82} \\
        Hindi & 56.97${\scriptstyle\pm}$\scriptsize{1.44} & 34.70${\scriptstyle\pm}$\scriptsize{4.71} & 10.62${\scriptstyle\pm}$\scriptsize{0.70} \\
        Japanese & 47.17${\scriptstyle\pm}$\scriptsize{1.92} & 5.7${\scriptstyle\pm}$\scriptsize{4.03} & 10.96${\scriptstyle\pm}$\scriptsize{0.91}  \\
        Korean & 46.67${\scriptstyle\pm}$\scriptsize{0.92} & 21.56${\scriptstyle\pm}$\scriptsize{7.62} & \textbf{15.03}${\scriptstyle\pm}$\scriptsize{0.81} \\
        \midrule
        All & 56.92${\scriptstyle\pm}$\scriptsize{1.12} & 55.58${\scriptstyle\pm}$\scriptsize{0.80} & 28.05${\scriptstyle\pm}$\scriptsize{1.52} \\
        \bottomrule
    \end{tabular}}
    \caption{Results on the EusIE dataset.}
    \label{tab:testB}
\end{table}

The results for argument extraction are significantly lower than for the other two tasks. The task is harder to transfer from language to language and might be severely affected by the under-sampling of the training set. Language-wise, the best-performing language is Korean, followed very closely by Turkish. English and Portuguese are also significantly better than the average. Spanish is a special case due to its marginal amount of training data. We had to consider the next smaller language because Spanish was too small for argument extraction. 

Finally, we can see that the \textit{All} model outperforms the rest of the languages on event detection and event argument extraction, but not on entity detection. We hypothesize that adding several languages is helpful when training datasets are small, but when there is enough data, it introduces noise. 



\subsection{Data scaling results}

We showed that the results on the event argument extraction are significantly lower. We hypothesized that this is due to the insufficient amount of the training data to learn properly the task. The fact that using all languages for training nearly doubles the best results for argument extraction in Table~\ref{tab:testB} is some evidence in this direction. 

\begin{figure}
    \centering
    \includegraphics[width = \linewidth]{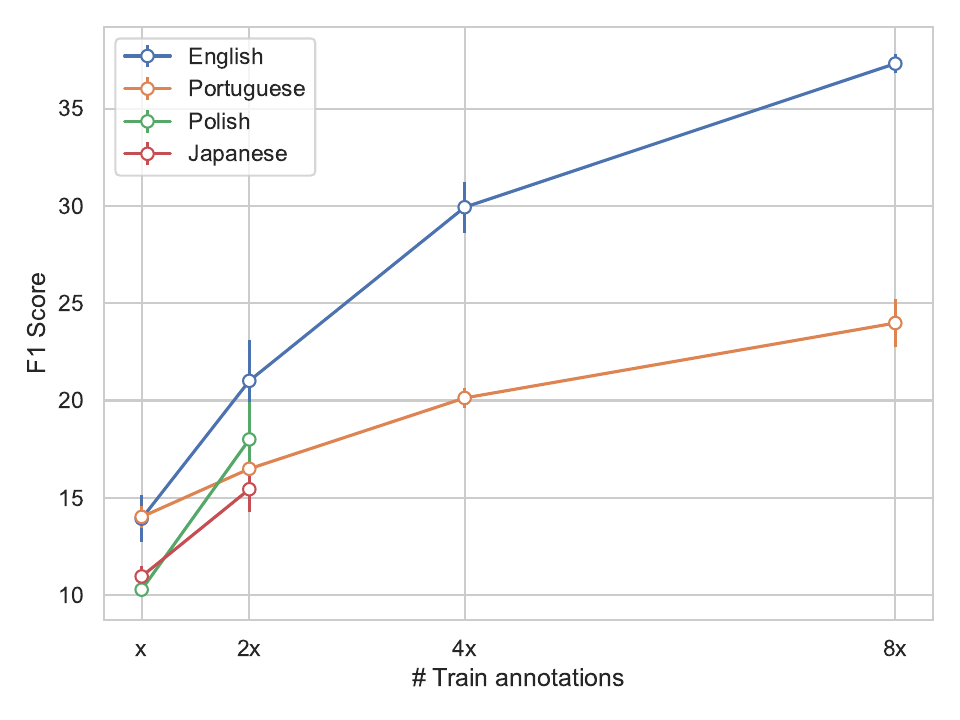}
    \caption{Train data scaling experiments on the event argument extraction task. $x=1,416$ instances.}
    \label{fig:data-scale}
\end{figure}

To study the effects of the training size in the cross-lingual setting, we trained new models with different amounts of data as shown in Figure~\ref{fig:data-scale}. Here, the $x$ denotes the initial training amount (1,416 instances), the same as used in Table~\ref{tab:testB}. We scaled the $x$ by 2, 4, and, 8 when possible.

Results in Figure~\ref{fig:data-scale} confirm our hypothesis: argument extraction is more challenging and requires more data to properly model the task. The figure shows how fast languages scale the performance with more data. Despite using fewer languages, we can see that languages scale at a different pace with the training data. For example, a comparison of English and Portuguese reveals that while both perform very similarly on the initial values, Portuguese scales much slower in the long term. 

Significantly English x8 outperforms the results attained with \textit{All} (in Table~\ref{tab:testB}) by a large margin.  Note that the amount of training data used is the same as the \textit{All} results but using only monolingual data. Together with the results for entities shown in the previous section, we can conclude that training size is a relevant factor in cross-lingual transfer and that mixing the data from all languages is beneficial only for smaller training sizes. 

\begin{table*}
    \centering
    \resizebox{1.\textwidth}{!}{\begin{tabular}{l|ccccc}
        \toprule
         Language & Morphology & Morphosyntactic Align. & Word Order & Script & Geographical Location \\
         \midrule
         English & Fusional & Nominative-Accusative & SVO & Latin & West Europe* \\
         Spanish & Fusional & Nominative-Accusative & SVO & Latin & West Europe* \\
         Portuguese & Fusional & Nominative-Accusative & SVO & Latin & West Europe* \\
         Polish & Fusional & Nominative-Accusative & SVO & Latin & East Europe \\
         Turkish & Agglutinative & Nominative-Accusative & SOV & Latin & East Europe \\
         Hindi & Fusional & Split Ergative & SOV & Devanagari & India \\
         Japanese & Agglutinative & Nominative-Accusative & SOV & Kanji \& Kana & East Asia \\
         Korean & Agglutinative & Nominative-Accusative & SOV & Hangul & East Asia \\
         \midrule
         Basque & Agglutinative & Ergative-Absolutive & SOV* & Latin & West Europe \\
         \bottomrule
    \end{tabular}}
    \caption{Language typology features. * indicates that the values are simplified, see main text for details.}
    \label{tab:languages}
\end{table*}

\section{Analysis according to Language Typology}


From the results on EusIE, we can draw two main conclusions: (1) There is no 
dominant language across tasks, and, (2) A language that is effective in a particular cross-lingual task does not guarantee that it will be good for another. For further understanding, we run an analysis on the following hypotheses: 

\begin{enumerate}
    \item Similar languages should benefit more from cross-lingual transfer.
    \item Different tasks require different skills. The tasks of detecting entities and events require more lexical knowledge, whereas extracting arguments requires a more structural understanding of the text.
\end{enumerate}

Focusing only on Basque as the target language would be limiting. So we decided to carry out the same experiments we did for Basque, but in this case, running the cross-lingual experiments for all the possible language pairs. We exclude the combinations that include the same source and target languages. 


\subsection{Language categorization}
\begin{figure*}
    \centering
    \includegraphics[width = \linewidth]{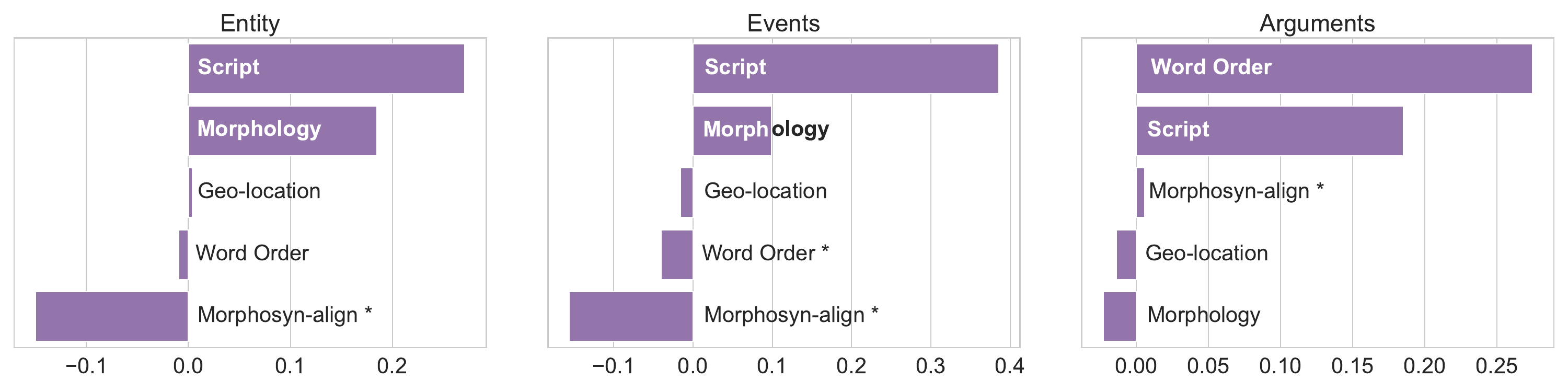}
    \caption{Estimated feature impact for each task. * indicates the result is not statistically significant (\texttt{p\_value >= 0.01}).}
    \label{fig:feature-analysis}
\end{figure*}

To validate our hypotheses, we first defined a set of linguistic typology features that help categorize the languages in our dataset.  We selected the features that could be relevant for cross-lingual transfer. 
Table~\ref{tab:languages} summarizes our categorization\footnote{The categorization was based on Wikipedia.}.

\paragraph{Morphology} refers to the study of words and how they are formed. We categorized each language into \textit{Agglutinative} or \textit{Fusional} categories. Our initial guess is that languages with similar morphology should perform better on tasks requiring more lexical knowledge (Entities and Events).

\paragraph{Morphosyntactic alignment} refers to the study of the relationship between different arguments of verbs. We categorized the languages into \textit{Nominative-Accusative}, \textit{Ergative-Absolutive}, and \textit{Split-Ergative}. As it is directly related to how the event arguments are marked in the sentence, we guess that languages with similar morphosyntactic alignment will perform better.  Note that most languages are categorized as \textit{Nominative-Accusative}, making it difficult to measure the effects of this feature.

\paragraph{Word order} refers to the order of the syntactic constituents of a language. We categorized the languages into \textit{Subject-Object-Verb} (SOV) or \textit{Subject-Verb-Object} (SVO) categories. An important consideration is that Basque usually follows the SOV order~\cite{de_rijk}, however, but also allows other orders depending on pragmatic factors~\cite{Laka_1996}. The word order has a significant impact when defining the different roles each part of the sentence has with respect to the verb, and therefore to an event. We guess that it would positively affect the event argument extraction task.

\paragraph{Script} refers to how the language is written. This is particularly important because it affects directly the tokenizer of the model, and therefore, how the token is defined and how the information is stored in the model 
we define very broad categories, in which we distinguish languages into two main groups: 
\textit{Latin} based and \textit{non-Latin} based. Our guess is that the script should affect all the tasks, as it is an important feature that impacts directly the model architecture. Although, with the same script, words from different languages might be better or worse represented depending on how the tokenizer was constructed.

\paragraph{Geographical location} refers to where the language is spoken. It is impossible to determine a concrete geographical location for a language, as nowadays a lot of languages are spoken all around the world, particularly English, Spanish, and Portuguese. However, for simplification purposes, we will consider them as languages spoken mainly in the west of Europe. The location of a language has a great cultural impact, and therefore, different entities appear more frequently in text on one language than another. This feature, however, is highly correlated with the script feature, as geographically adjacent languages tend to have similar scripts. We think this feature may have a greater impact on lexical tasks. 

\subsection{Results of the analysis}

To analyze the effect of each typological feature, we ran the same cross-lingual experiments as we did with Basque, but in this case, running all possible source-target language combinations. The results are shown in Table~\ref{tab:all_results} (in Appendix~\ref{ap:results}). Based on these results we run multiple regression analysis for each task separately and use the resulting coefficients to measure the relative contribution of each linguistic feature described above.

We prepare the data in order to correctly perform the analysis. First, we normalized the results across target languages as we noticed that the values on each target language ranged very differently (even across the same task). That is, we transform using Min-Max scaling the F1 scores of each task and target language into values between 0 and 1. Regarding input variables, we generate features  that indicate whether the value of the given source $f_i^{source}$ and target $f_i^{target}$ languages are the same: 

\begin{equation}
    f_i = \left\{
    \begin{array}{lr}
        1 & f_i^{source} = f_i^{target} \\
        0 & \, \textrm{\small otherwise} \\
    \end{array}
\right.
\end{equation}

We formalize the linear regression analysis as shown in Equation~\ref{eq:reg}, where $s$ is the normalized F1 score for each language and task, and $w_i$ is the coefficient that measures the relative contribution of feature $i$. As additional information, Figure~\ref{fig:boxplots} (in Appendix~\ref{ap:results}) shows the distribution of normalized F1 scores for each linguistic feature in isolation. 

\begin{equation}
    s = w_o + \sum_{i}^{|F|} w_i \cdot f_i
    \label{eq:reg}
\end{equation}

Figure~\ref{fig:feature-analysis} shows the contribution of each linguistic feature to performance in each task. \textit{Script} turns out to be the most relevant feature across tasks. \textit{Morphology} is an important feature in entity and event detection, where lexical information could play a significant role, as we hypothesized. The importance of these two features might be due to the fact that languages with the same script and morphology share more tokens in the language model vocabulary. On the other hand, \textit{Word Order}, as we hypothesized, affects significantly the argument extraction task.  Surprisingly,  \textit{Geo-location} is not relevant when transferring knowledge from one language to another, it is well-known that geographically close languages share many lexical entries. Therefore, the relevance of geo-location is low probably because most of the correlation is explained with \textit{Script}.  As expected we did not find any correlation for \textit{Morphosyntactic alignment}, probably because all languages but one shared the same feature. 

All in all the analysis shows that sharing the script is a key factor in all three tasks and that sharing the script might overshadow the relevance of sharing the geographical location. As we hypothesized, morphology is relevant for the two tasks which are more lexical (entity and event extraction), while word order turns out to be relevant for the more syntactic argument extraction. 



\section{Conclusions}
In this paper we explore the contribution of different training languages when transferring into other languages, presenting a set of exhaustive experiments on three Event Extraction tasks and eight languages with different language typological features. In a first experiment with Basque as a target, we see that there is no clear pattern. In a subsequent experiment, we performed a typologically motivated correlation analysis over all the language combinations and concluded that transfer quality does correlate with some linguistic features, which change depending on the task. For entity and trigger identification sharing the script and morphological typology between source and target languages are the two most relevant features. In argument extraction sharing word order and script play the most relevant roles. In addition, we show that source languages scale differently as we increase training data. Finally, we present the first Basque Event Extraction evaluation benchmark, which was used alongside the MEE dataset~\citet{pouran-ben-veyseh-etal-2022-mee} in all experiments.


For the future, we would like to extend the analysis to more tasks and languages, as well as taking into account other alternative typological features. A better understanding of the interactions between typology and cross-lingual transfer opens a new research avenue that can be beneficial for low-resource languages.

\section*{Acknowledgements}

This work has been partially supported by 
the Basque Government (Research group funding IT-1805-22, IKER-GAITU and ICL4LANG Grant no. KK-2023/00094). We are also thankful to a MCIN/AEI/10.13039/501100011033 DeepKnowledge (PID2021-127777OB-C21) and FEDER, EU. This work has been also funded by The EFA104/01 LINGUATEC-IA project “Cross-border network of technological cooperation in artificial intelligence applied to language for the construction of a trans-Pyrenean linguistic infrastructure” is 65\% co-financed by the European Regional Development Fund (ERDF) through the Interreg V-A Program Spain-France-Andorra (POCTEFA 2021-2027).

\section*{Bibliographical References}\label{sec:reference}

\bibliographystyle{lrec-coling2024-natbib}
\bibliography{lrec-coling2024-example, anthology}

\section*{Language Resource References}
\label{lr:ref}
\bibliographystylelanguageresource{lrec-coling2024-natbib}
\bibliographylanguageresource{languageresource}

\newpage
\appendix

\section{Additional results} \label{ap:results}

\begin{table*}
    \centering
    \begin{tabular}{l|rrr|rrr}
        \toprule
        & \multicolumn{3}{c|}{Ours} & \multicolumn{3}{c}{MEE} \\ 
        Languages & \multicolumn{1}{c}{Entities} & \multicolumn{1}{c}{Events} & \multicolumn{1}{c|}{Arguments} & \multicolumn{1}{c}{Entities} & \multicolumn{1}{c}{Events} & \multicolumn{1}{c}{Arguments} \\
        \midrule
        English & 80.48${\scriptstyle\pm}$\scriptsize{0.19} & 78.47${\scriptstyle\pm}$\scriptsize{0.33} & 63.60${\scriptstyle\pm}$\scriptsize{0.09} & 70.22 & 71.28 & 66.34  \\
        Spanish & 84.56${\scriptstyle\pm}$\scriptsize{0.38} & 63.86${\scriptstyle\pm}$\scriptsize{1.20} & 40.45${\scriptstyle\pm}$\scriptsize{1.91} & 70.33 & 64.32 & 61.12  \\
        Portuguese & 80.42${\scriptstyle\pm}$\scriptsize{0.32} & 61.79${\scriptstyle\pm}$\scriptsize{0.63} & 68.18${\scriptstyle\pm}$\scriptsize{0.99} & 70.39 & 71.88 & 71.75\\
        Polish & 81.00${\scriptstyle\pm}$\scriptsize{0.40} & 69.09${\scriptstyle\pm}$\scriptsize{0.91} & 76.25${\scriptstyle\pm}$\scriptsize{1.26} & 69.14 & 60.45 & 61.23  \\
        Turkish & 70.83${\scriptstyle\pm}$\scriptsize{0.06} & 56.62${\scriptstyle\pm}$\scriptsize{0.43} & 24.08${\scriptstyle\pm}$\scriptsize{1.66}  & 76.13 & 67.18 & 55.78  \\
        Hindi & 76.00${\scriptstyle\pm}$\scriptsize{0.41} & 48.21${\scriptstyle\pm}$\scriptsize{2.54} & 45.31${\scriptstyle\pm}$\scriptsize{1.55} & 65.14 & 59.34 & 58.22  \\
        Japanese & 47.43${\scriptstyle\pm}$\scriptsize{0.20} & 35.74${\scriptstyle\pm}$\scriptsize{1.92} & 52.93${\scriptstyle\pm}$\scriptsize{1.26}  & 71.34 & 67.77 & 69.19 \\
        Korean & 71.31${\scriptstyle\pm}$\scriptsize{0.78} & 45.30${\scriptstyle\pm}$\scriptsize{0.49} & 34.71${\scriptstyle\pm}$\scriptsize{3.06} & 59.13 & 62.34 & 69.70 \\
        \midrule
        All & 78.63${\scriptstyle\pm}$\scriptsize{0.17} & 68.01${\scriptstyle\pm}$\scriptsize{0.27} & 59.74${\scriptstyle\pm}$\scriptsize{0.99} & \multicolumn{1}{c}{-} & \multicolumn{1}{c}{-} & \multicolumn{1}{c}{-} \\
        \bottomrule
    \end{tabular} 
    \caption{Comparison between our system and that of the original MEE paper.}
    \label{tab:comparison-mee}
\end{table*}

\begin{table*}
    \centering
    \resizebox{1.\textwidth}{!}{\begin{tabular}{lrrrrrrrr}
\toprule
\multicolumn{9}{c}{Entities} \\
Source lang.  & English & Spanish & Portuguese & Polish & Turkish & Hindi & Japanese & Korean \\
\midrule
English & \multicolumn{1}{c}{-} & \textbf{76.00}${\scriptstyle\pm}$\scriptsize{0.00} & 68.67${\scriptstyle\pm}$\scriptsize{0.58} & 66.67${\scriptstyle\pm}$\scriptsize{0.58} & 57.33${\scriptstyle\pm}$\scriptsize{1.53} & 67.00${\scriptstyle\pm}$\scriptsize{1.00} & 29.33${\scriptstyle\pm}$\scriptsize{1.15} & 46.00${\scriptstyle\pm}$\scriptsize{1.00} \\
Spanish & \textbf{72.33}${\scriptstyle\pm}$\scriptsize{1.15} & \multicolumn{1}{c}{-} & \textbf{72.67}${\scriptstyle\pm}$\scriptsize{0.58} & \textbf{69.67}${\scriptstyle\pm}$\scriptsize{0.58} & 59.67${\scriptstyle\pm}$\scriptsize{0.58} & \textbf{68.67}${\scriptstyle\pm}$\scriptsize{0.58} & 27.67${\scriptstyle\pm}$\scriptsize{1.53} & 47.33${\scriptstyle\pm}$\scriptsize{0.58} \\
Portuguese & 66.33${\scriptstyle\pm}$\scriptsize{0.58} & 71.00${\scriptstyle\pm}$\scriptsize{0.00} & \multicolumn{1}{c}{-} & 58.33${\scriptstyle\pm}$\scriptsize{0.58} & 50.67${\scriptstyle\pm}$\scriptsize{0.58} & 60.00${\scriptstyle\pm}$\scriptsize{1.00} & 27.00${\scriptstyle\pm}$\scriptsize{2.00} & 40.33${\scriptstyle\pm}$\scriptsize{0.58} \\
Polish & 69.00${\scriptstyle\pm}$\scriptsize{0.00} & 73.33${\scriptstyle\pm}$\scriptsize{0.58} & 67.00${\scriptstyle\pm}$\scriptsize{1.73} & \multicolumn{1}{c}{-} & 59.33${\scriptstyle\pm}$\scriptsize{0.58} & 68.33${\scriptstyle\pm}$\scriptsize{0.58} & 31.00${\scriptstyle\pm}$\scriptsize{2.00} & 48.00${\scriptstyle\pm}$\scriptsize{1.73} \\
Turkish & 71.33${\scriptstyle\pm}$\scriptsize{0.58} & 70.33${\scriptstyle\pm}$\scriptsize{1.53} & 63.67${\scriptstyle\pm}$\scriptsize{1.53} & 66.33${\scriptstyle\pm}$\scriptsize{0.58} & \multicolumn{1}{c}{-} & 70.00${\scriptstyle\pm}$\scriptsize{0.00} & 33.00${\scriptstyle\pm}$\scriptsize{2.65} & \textbf{51.33}${\scriptstyle\pm}$\scriptsize{1.15} \\
Hindi & 67.33${\scriptstyle\pm}$\scriptsize{1.53} & 70.33${\scriptstyle\pm}$\scriptsize{1.53} & 62.67${\scriptstyle\pm}$\scriptsize{2.08} & 65.67${\scriptstyle\pm}$\scriptsize{1.15} & 58.33${\scriptstyle\pm}$\scriptsize{0.58} & \multicolumn{1}{c}{-} & 35.00${\scriptstyle\pm}$\scriptsize{0.00} & 48.00${\scriptstyle\pm}$\scriptsize{0.00} \\
Japanese & 40.33${\scriptstyle\pm}$\scriptsize{5.69} & 43.33${\scriptstyle\pm}$\scriptsize{6.51} & 41.00${\scriptstyle\pm}$\scriptsize{2.00} & 48.00${\scriptstyle\pm}$\scriptsize{5.29} & 48.00${\scriptstyle\pm}$\scriptsize{2.65} & 44.67${\scriptstyle\pm}$\scriptsize{5.03} & \multicolumn{1}{c}{-} & 41.67${\scriptstyle\pm}$\scriptsize{1.53} \\
Korean & 60.00${\scriptstyle\pm}$\scriptsize{1.00} & 58.00${\scriptstyle\pm}$\scriptsize{1.73} & 53.67${\scriptstyle\pm}$\scriptsize{0.58} & 60.67${\scriptstyle\pm}$\scriptsize{0.58} & \textbf{63.33}${\scriptstyle\pm}$\scriptsize{0.58} & 59.33${\scriptstyle\pm}$\scriptsize{0.58} & \textbf{37.00}${\scriptstyle\pm}$\scriptsize{1.00} & \multicolumn{1}{c}{-} \\
\midrule
\multicolumn{9}{c}{Events} \\
Source lang.  & English & Spanish & Portuguese & Polish & Turkish & Hindi & Japanese & Korean \\
\midrule
English & \multicolumn{1}{c}{-} & \textbf{52.33}${\scriptstyle\pm}$\scriptsize{0.58} & 37.67${\scriptstyle\pm}$\scriptsize{1.15} & \textbf{50.67}${\scriptstyle\pm}$\scriptsize{5.77} & \textbf{48.67}${\scriptstyle\pm}$\scriptsize{3.21} & 43.33${\scriptstyle\pm}$\scriptsize{1.53} & 1.33${\scriptstyle\pm}$\scriptsize{2.31} & \textbf{35.00}${\scriptstyle\pm}$\scriptsize{2.65} \\
Spanish & 53.00${\scriptstyle\pm}$\scriptsize{2.65} & \multicolumn{1}{c}{-} & \textbf{39.00}${\scriptstyle\pm}$\scriptsize{1.00} & 40.67${\scriptstyle\pm}$\scriptsize{3.21} & 48.00${\scriptstyle\pm}$\scriptsize{1.00} & 45.33${\scriptstyle\pm}$\scriptsize{2.52} & 0.33${\scriptstyle\pm}$\scriptsize{0.58} & 32.67${\scriptstyle\pm}$\scriptsize{2.89} \\
Portuguese & 50.00${\scriptstyle\pm}$\scriptsize{0.00} & 37.00${\scriptstyle\pm}$\scriptsize{4.36} & \multicolumn{1}{c}{-} & 46.33${\scriptstyle\pm}$\scriptsize{3.51} & 37.00${\scriptstyle\pm}$\scriptsize{2.00} & 34.67${\scriptstyle\pm}$\scriptsize{0.58} & 0.00${\scriptstyle\pm}$\scriptsize{0.00} & 27.67${\scriptstyle\pm}$\scriptsize{1.15} \\
Polish & \textbf{61.67}${\scriptstyle\pm}$\scriptsize{0.58} & 49.00${\scriptstyle\pm}$\scriptsize{0.00} & 29.33${\scriptstyle\pm}$\scriptsize{6.11} & \multicolumn{1}{c}{-} & 45.00${\scriptstyle\pm}$\scriptsize{2.00} & 38.00${\scriptstyle\pm}$\scriptsize{5.57} & 0.00${\scriptstyle\pm}$\scriptsize{0.00} & 29.33${\scriptstyle\pm}$\scriptsize{4.04} \\
Turkish & 51.67${\scriptstyle\pm}$\scriptsize{11.1} & 46.67${\scriptstyle\pm}$\scriptsize{4.04} & 36.00${\scriptstyle\pm}$\scriptsize{1.73} & 39.00${\scriptstyle\pm}$\scriptsize{14.9} & \multicolumn{1}{c}{-} & \textbf{47.00}${\scriptstyle\pm}$\scriptsize{1.73} & 1.33${\scriptstyle\pm}$\scriptsize{1.53} & 33.33${\scriptstyle\pm}$\scriptsize{2.31} \\
Hindi & 56.00${\scriptstyle\pm}$\scriptsize{4.58} & 43.33${\scriptstyle\pm}$\scriptsize{6.66} & 18.33${\scriptstyle\pm}$\scriptsize{5.51} & 31.33${\scriptstyle\pm}$\scriptsize{11.5} & 42.00${\scriptstyle\pm}$\scriptsize{1.00} & \multicolumn{1}{c}{-} & 3.67${\scriptstyle\pm}$\scriptsize{2.08} & 30.00${\scriptstyle\pm}$\scriptsize{3.46} \\
Japanese & 9.00${\scriptstyle\pm}$\scriptsize{11.3} & 7.33${\scriptstyle\pm}$\scriptsize{10.9} & 7.00${\scriptstyle\pm}$\scriptsize{7.55} & 23.67${\scriptstyle\pm}$\scriptsize{8.39} & 19.33${\scriptstyle\pm}$\scriptsize{6.35} & 7.67${\scriptstyle\pm}$\scriptsize{8.08} & \multicolumn{1}{c}{-} & 24.33${\scriptstyle\pm}$\scriptsize{8.50} \\
Korean & 41.00${\scriptstyle\pm}$\scriptsize{5.00} & 21.67${\scriptstyle\pm}$\scriptsize{12.6} & 29.33${\scriptstyle\pm}$\scriptsize{5.51} & 28.67${\scriptstyle\pm}$\scriptsize{7.02} & 30.67${\scriptstyle\pm}$\scriptsize{1.15} & 31.33${\scriptstyle\pm}$\scriptsize{6.11} & \textbf{6.00}${\scriptstyle\pm}$\scriptsize{3.00} & \multicolumn{1}{c}{-} \\
\midrule
\multicolumn{9}{c}{Arguments} \\
Source lang.  & English & Spanish & Portuguese & Polish & Turkish & Hindi & Japanese & Korean \\
\midrule
English & \multicolumn{1}{c}{-} & \textbf{20.33}${\scriptstyle\pm}$\scriptsize{2.08} & \textbf{43.33}${\scriptstyle\pm}$\scriptsize{1.15} & 40.67${\scriptstyle\pm}$\scriptsize{1.53} & \textbf{14.67}${\scriptstyle\pm}$\scriptsize{3.21} & 15.33${\scriptstyle\pm}$\scriptsize{3.06} & 14.33${\scriptstyle\pm}$\scriptsize{1.53} & 14.00${\scriptstyle\pm}$\scriptsize{2.65} \\
Spanish & 13.00${\scriptstyle\pm}$\scriptsize{2.65} & \multicolumn{1}{c}{-} & 25.67${\scriptstyle\pm}$\scriptsize{2.31} & 14.00${\scriptstyle\pm}$\scriptsize{3.46} & 3.00${\scriptstyle\pm}$\scriptsize{3.46} & 8.00${\scriptstyle\pm}$\scriptsize{2.00} & 0.00${\scriptstyle\pm}$\scriptsize{0.00} & 2.00${\scriptstyle\pm}$\scriptsize{0.00} \\
Portuguese & \textbf{37.00}${\scriptstyle\pm}$\scriptsize{1.73} & 17.00${\scriptstyle\pm}$\scriptsize{2.65} & \multicolumn{1}{c}{-} & \textbf{42.00}${\scriptstyle\pm}$\scriptsize{2.00} & 13.67${\scriptstyle\pm}$\scriptsize{2.52} & 16.00${\scriptstyle\pm}$\scriptsize{1.73} & 19.33${\scriptstyle\pm}$\scriptsize{1.15} & 12.00${\scriptstyle\pm}$\scriptsize{1.73} \\
Polish & 31.00${\scriptstyle\pm}$\scriptsize{1.73} & 15.00${\scriptstyle\pm}$\scriptsize{3.00} & 36.00${\scriptstyle\pm}$\scriptsize{2.00} & \multicolumn{1}{c}{-} & 14.33${\scriptstyle\pm}$\scriptsize{1.53} & 14.33${\scriptstyle\pm}$\scriptsize{2.89} & 16.67${\scriptstyle\pm}$\scriptsize{2.08} & 14.00${\scriptstyle\pm}$\scriptsize{4.36} \\
Turkish & 22.33${\scriptstyle\pm}$\scriptsize{0.58} & 18.33${\scriptstyle\pm}$\scriptsize{2.52} & 31.67${\scriptstyle\pm}$\scriptsize{0.58} & 33.67${\scriptstyle\pm}$\scriptsize{1.15} & \multicolumn{1}{c}{-} & \textbf{29.33}${\scriptstyle\pm}$\scriptsize{1.15} & \textbf{20.00}${\scriptstyle\pm}$\scriptsize{2.65} & 13.33${\scriptstyle\pm}$\scriptsize{2.31} \\
Hindi & 14.33${\scriptstyle\pm}$\scriptsize{1.15} & 9.33${\scriptstyle\pm}$\scriptsize{4.04} & 19.00${\scriptstyle\pm}$\scriptsize{1.00} & 23.67${\scriptstyle\pm}$\scriptsize{1.53} & 12.33${\scriptstyle\pm}$\scriptsize{0.58} & \multicolumn{1}{c}{-} & 17.33${\scriptstyle\pm}$\scriptsize{0.58} & 10.33${\scriptstyle\pm}$\scriptsize{2.31} \\
Japanese & 12.67${\scriptstyle\pm}$\scriptsize{0.58} & 3.67${\scriptstyle\pm}$\scriptsize{0.58} & 14.00${\scriptstyle\pm}$\scriptsize{2.65} & 12.00${\scriptstyle\pm}$\scriptsize{1.00} & 11.33${\scriptstyle\pm}$\scriptsize{3.21} & 13.67${\scriptstyle\pm}$\scriptsize{1.53} & \multicolumn{1}{c}{-} & \textbf{20.33}${\scriptstyle\pm}$\scriptsize{3.06} \\
Korean & 14.00${\scriptstyle\pm}$\scriptsize{3.46} & 4.33${\scriptstyle\pm}$\scriptsize{1.15} & 16.00${\scriptstyle\pm}$\scriptsize{1.00} & 18.00${\scriptstyle\pm}$\scriptsize{3.61} & 7.67${\scriptstyle\pm}$\scriptsize{2.31} & 17.00${\scriptstyle\pm}$\scriptsize{1.00} & 17.33${\scriptstyle\pm}$\scriptsize{2.31} & \multicolumn{1}{c}{-} \\
\bottomrule
\end{tabular}}
    \caption{Results for every combination of source-target languages excluding Basque. Rows indicate the source language and columns the target. For every combination 3 different runs were done in order to compute the mean and standard deviation.}
    \label{tab:all_results}
\end{table*}

\begin{figure*}
    \centering
    \includegraphics[width = \linewidth]{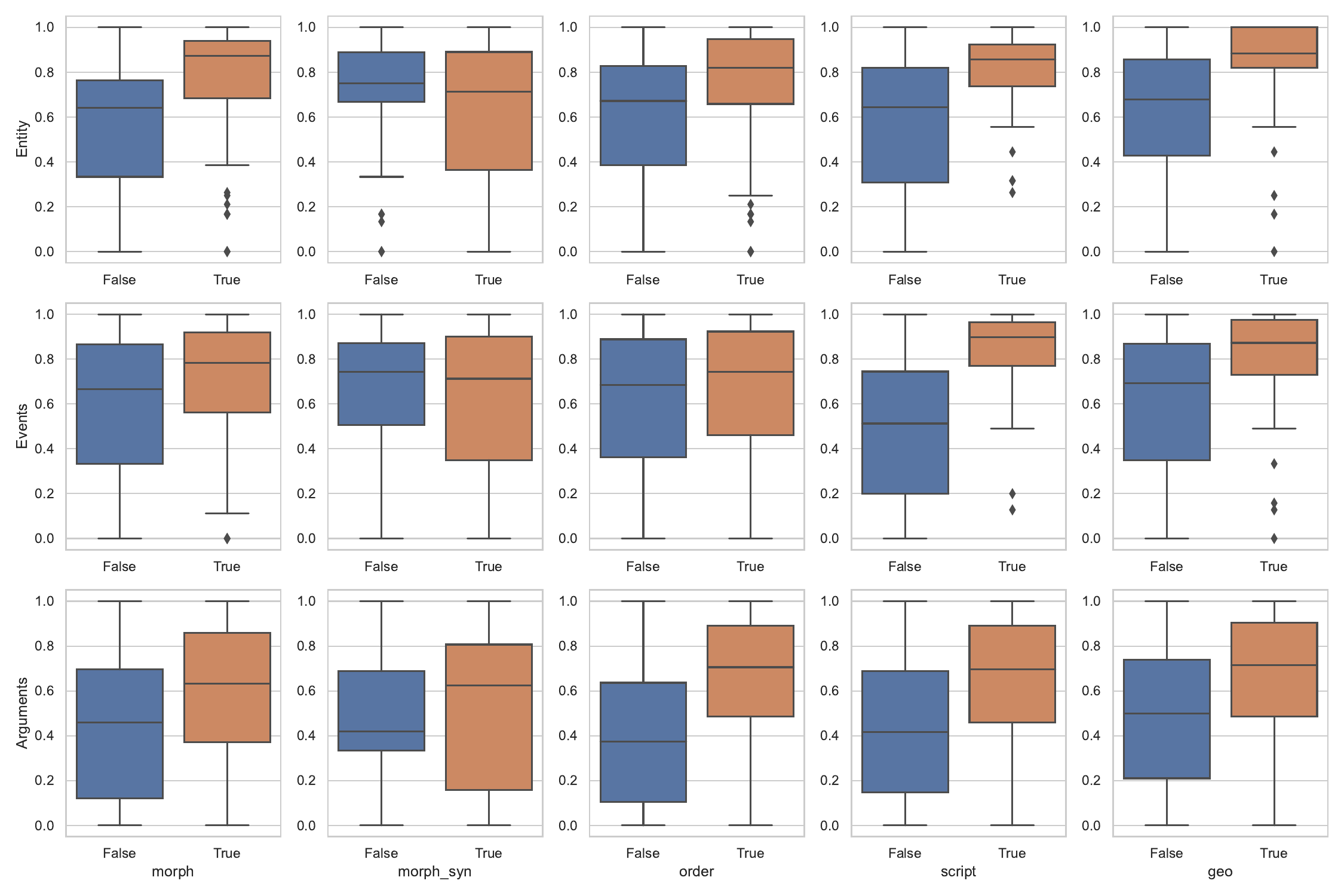}
    \caption{Boxplots of each task and features. The label "True" indicates that source and target languages sharing the same value for a given feature.}
    \label{fig:boxplots}
\end{figure*}

\end{document}